\documentclass[11pt,a4paper]{article}

\usepackage[T1]{fontenc}
\usepackage[utf8]{inputenc}
\usepackage{lmodern}
\usepackage{microtype}
\usepackage[margin=1in]{geometry}
\usepackage{amsmath,amssymb,amsthm}
\usepackage{graphicx}
\usepackage{booktabs}
\usepackage{tikz}
\usepackage{xcolor}
\usepackage{authblk}
\usepackage[numbers,sort&compress]{natbib}
\usepackage[hidelinks]{hyperref}

\title{YingLan: A Multimodal Ocean Foundation Model for Unified Representation Learning from Heterogeneous Observations}

\author[1]{Sen Jia}
\author[1]{Siqi Wei}
\author[1]{Junyu Wang}
\author[1]{Zijun Lv}
\author[1]{Pan Liu}
\author[1]{Meng Xu}
\author[1]{Xiyou Fu}
\author[1]{Yuxi Sun}
\author[1]{Qiang Liu}
\author[1]{Shuyu Zhang\textsuperscript{*}}
\author[2]{Lin Mu\textsuperscript{*}}

\affil[1]{College of Computer Science and Software Engineering, Shenzhen University}

\affil[2]{College of Life Sciences and Oceanography, Shenzhen University}

\affil[1*,\,2*]{\small
Corresponding authors:
Shuyu Zhang (\texttt{shuyu-zhang@szu.edu.cn});
Lin Mu (\texttt{mulin@szu.edu.cn})
}

\date{} 

\begin{document}
\maketitle
\begin{abstract}
Multimodal ocean modeling remains challenging due to
substantial heterogeneity in data structures, physical
semantics, and spatiotemporal characteristics across
satellite imagery, gridded oceanographic fields, and
buoy observations. Existing approaches largely focus
on task-specific ocean modeling, while learning
transferable representations from heterogeneous
observations remains an open problem.
To address this challenge, we propose \textbf{YingLan},
a multimodal ocean foundation model that integrates
physics-attribute-conditioned dynamic encoding,
Spatiotemporal Increment Coupling (STIC), and
ocean-state evolution-aware pretraining.
Specifically, the dynamic encoding mechanism generates
adaptive input projections conditioned on physical
semantics and observation characteristics, enabling
flexible representation of heterogeneous ocean variables.
STIC models multiscale spatiotemporal dependencies
through cross-scale interactions of latent temporal
increments. The pretraining strategy further combines
masked observation reconstruction with temporal
evolution objectives to facilitate the learning of
coherent ocean-state representations across different
prediction horizons.
YingLan is implemented using MindSpore and trained
on Huawei Ascend 910B NPUs, demonstrating its
compatibility with domestic AI computing infrastructure.
In a seven-day sea surface salinity (SSS) forecasting
task conducted in the Shenzhen coastal waters,
YingLan achieves an average RMSE of 0.687508 PSU,
corresponding to relative error reductions of
0.58\%, 5.31\%, and 12.42\% compared with LangYa,
XiHe, and GloNet2, respectively.
These results provide initial evidence of the model's
effectiveness in regional ocean forecasting, while
broader cross-task and cross-regional evaluations
remain necessary to assess its generalization
capabilities.
\end{abstract}

\noindent\textbf{Keywords:}
Ocean foundation model;
Multimodal representation learning;
Indigenous deployment;
Downstream task adaptation;
Self-supervised pretraining

\section{Introduction}
\label{sec:introduction}
Accurate characterization and prediction of the marine environment are fundamental to oceanographic research and marine environmental monitoring, with important implications for marine ecosystem conservation, marine resource management, and ocean hazard early warning \cite{YUAN2023159741}. Advances in satellite remote sensing, ocean numerical modeling, and in situ observation technologies have expanded the spatial coverage, temporal resolution, and range of physical variables represented in multi-source ocean observation data, providing an essential basis for multiscale characterization and dynamic analysis of the marine environment \cite{10.3389/fmars.2024.1363863}. Specifically, satellite imagery captures the spatial distribution of ocean surface conditions, oceanographic products provide spatiotemporal information on variables such as temperature, salinity, and ocean currents, and in situ observations from buoys and other platforms record continuous variations in local marine conditions \cite{10.3389/fmars.2023.1085542}. These observation sources are complementary in terms of spatial coverage, temporal sampling, and physical information. However, substantial differences in data structure, variable semantics, and spatiotemporal scales pose challenges to joint modeling of heterogeneous observations and knowledge transfer across downstream tasks. Consequently, large-scale pretraining on heterogeneous ocean observations to learn generalizable representations with cross-modal adaptability and spatiotemporal generalization has emerged as a promising research direction in ocean intelligence \cite{11296896}.

In recent years, deep learning has made substantial progress in ocean environmental forecasting and the development of ocean foundation models. For spatiotemporal ocean modeling, XiHe employs local and global spatial interactions to perform high-resolution \cite{wang2024xihedatadrivenmodelglobal}, multivariate global ocean forecasting, whereas WenHai integrates air-sea flux computation with ocean-dynamics-informed network design, highlighting the importance of spatial dependency modeling and the incorporation of domain knowledge in ocean prediction \cite{cui2025wenhai}. In the development of foundation models, AI-GOMS explores a pretraining and downstream adaptation paradigm for multiple ocean-related tasks, demonstrating the potential of shared representations to support diverse applications \cite{xiong2023aigomslargeaidrivenglobal}. Meanwhile, ocean remote sensing research has increasingly investigated the joint use of complementary observations during pretraining. The Sentinel-3 ocean-colour foundation model jointly pretrains on ocean-colour reflectance and sea surface temperature through masked modeling and evaluates the resulting representations on ocean-colour-related tasks \cite{dawson2025sentinel3foundationmodelocean}. OceanMAE combines multispectral imagery with auxiliary ocean descriptors for self-supervised reconstruction and assesses the transferability of its learned representations on ocean segmentation and bathymetry regression \cite{stamer2026oceanmaefoundationmodelocean}. Despite these advances in ocean forecasting, multi-source information utilization, and domain-specific pretraining, differences in the physical attributes, spatiotemporal scales, and temporal dynamics of heterogeneous observations continue to pose challenges for unified representation learning and joint modeling. Effectively encoding heterogeneous ocean observations, characterizing dependencies across spatial scales and temporal horizons, and learning ocean-state evolution through pretraining remain important open research questions for multimodal ocean foundation models.

Despite the progress of existing studies, developing multimodal foundation models that jointly leverage satellite imagery, oceanographic fields, and buoy observations remains challenging in terms of heterogeneous representation learning, multiscale spatiotemporal modeling, and the learning of ocean-state dynamics. First, these observation types differ substantially in data structure, physical semantics, and sampling characteristics \cite{NEURIPS2023_f6ccbf94,cong2023satmaepretrainingtransformerstemporal}. Satellite imagery primarily captures the spatial distribution and textural characteristics of ocean surface conditions, oceanographic fields describe the spatial structures of variables such as temperature, salinity, and current velocity, while buoy observations record localized temporal variations at specific locations. Fixed input representations may not adequately accommodate these differences in observational attributes \cite{NEURIPS2023_11822e84,nedungadi2024mmearthexploringmultimodalpretext}, potentially limiting the effective utilization of heterogeneous information within a shared representation space. Second, marine environments exhibit multiscale spatiotemporal variability, in which local spatial structures, broader environmental contexts, and temporal variations over different horizons are interrelated. Single-scale feature extraction may therefore be insufficient to characterize these complex dependencies. Furthermore, pretraining based solely on reconstructing observed data does not necessarily encourage models to learn ocean-state evolution patterns that are predictive of future conditions \cite{bodnar2025foundation}. Consequently, further research is needed to develop transferable multimodal ocean foundation models that effectively represent heterogeneous observations, capture multiscale spatiotemporal dependencies, and learn the temporal evolution of ocean states through pretraining.

To address these challenges, we propose \textbf{YingLan}, a multimodal
ocean foundation model for unified representation learning from
heterogeneous ocean observations, including satellite imagery, gridded
oceanographic fields, and buoy time series. YingLan integrates
physics-attribute-conditioned dynamic encoding, multiscale spatiotemporal
interaction, and ocean-state evolution-aware pretraining to learn
transferable representations of ocean states across heterogeneous
observation types and temporal scales. The main contributions of this
work are summarized as follows:

\begin{itemize}
    \item \textbf{A multimodal foundation model for heterogeneous ocean
    observations.}
    We develop a unified modeling framework that accommodates satellite
    imagery, gridded oceanographic fields, and buoy time series, providing
    a shared representation space for heterogeneous ocean observations
    and supporting adaptation to diverse regional ocean monitoring and
    forecasting tasks.

    \item \textbf{Physics-attribute-conditioned dynamic encoding.}
    We introduce a dynamic encoding mechanism that jointly incorporates
    physical semantics and observation conditions into input projection
    parameter generation. By distinguishing variable-specific physical
    attributes from observation-specific characteristics, the proposed
    mechanism enables adaptive encoding of heterogeneous ocean
    observations.

    \item \textbf{Multiscale spatiotemporal interaction and evolution-aware
    pretraining.}
    We propose a Spatiotemporal Increment Coupling (STIC) module that
    facilitates cross-scale information exchange through latent temporal
    increments, enabling interactions between localized ocean structures
    and large-scale environmental contexts. We further develop an
    ocean-state evolution-aware pretraining strategy that jointly
    optimizes masked observation reconstruction, horizon-conditioned
    latent increment prediction, and temporal compositional consistency
    to learn temporally coherent representations of ocean-state dynamics.

    \item \textbf{Evaluation of regional transferability and
    generalization.}
    We investigate the transferability of YingLan in regional ocean
    applications, using Shenzhen coastal waters as a case study.
    Through downstream forecasting and, where supported by the
    experimental design, few-shot adaptation, cross-temporal evaluation,
    and cross-region transfer, we assess the model's predictive
    performance and generalization under varying data availability
    and spatiotemporal conditions.
\end{itemize}

\section{Related Work}
\label{sec:related}

\subsection{Ocean Foundation Models and Pretraining Paradigms}
Existing deep learning studies on ocean modeling have primarily focused on ocean state estimation and evolution forecasting, aiming to characterize ocean physical processes and their spatiotemporal variability through data-driven approaches. Eeltink et al. \cite{eeltink2022nonlinear} proposed a hybrid approach that integrates recurrent neural networks with a nonlinear wave evolution model, learning data-driven correction terms to represent wave-breaking processes and illustrating the integration of physical models with neural networks for ocean dynamics modeling. Ham et al. \cite{ham2024partial} proposed a global ocean data assimilation framework based on partial convolutions and generative adversarial learning, reconstructing ocean states by incorporating observational information and improving the handling of incomplete ocean observations. Furthermore, Cui et al. \cite{cui2025wenhai} introduced WenHai, which employs a Swin Transformer-based architecture to model multivariate ocean state evolution and incorporates air–sea flux information and multi-step forecasting strategies. These studies advance ocean deep learning from complementary perspectives, including physical process modeling, ocean state reconstruction, and spatiotemporal forecasting. Nevertheless, their training objectives remain primarily tied to specific ocean variables and predefined estimation or forecasting tasks, motivating the exploration of pretrained representations that can be transferred across diverse ocean applications.

Recent ocean foundation models have explored diverse pretraining strategies tailored to different types of ocean data. Xiong et al. \cite{xiong2023aigomslargeaidrivenglobal} proposed AI-GOMS, which learns ocean-state representations from HYCOM reanalysis fields through Fourier-based masked autoencoding and supports downstream adaptation across ocean modeling tasks. Glaser et al. \cite{glaser2025wvnet} introduced WV-Net, using contrastive self-supervised learning on large-scale Sentinel-1 SAR imagery to obtain transferable representations for ocean remote sensing. Complementarily, Bi et al. \cite{bi2024oceangpt} developed OceanGPT through ocean-science corpus pretraining and domain-specific instruction tuning, while Xue et al. \cite{xue2026oceanpile} introduced OceanPile, a multimodal ocean-science corpus designed to support foundation model training across marine imagery, sonar data, and scientific text. These studies span physical-field pretraining, image-based representation learning, and language-oriented ocean knowledge modeling, illustrating the diversity of data and objectives in ocean-domain pretraining. However, their learning frameworks address different data modalities and task requirements, leaving the unified representation of heterogeneous ocean observations an important direction for further investigation.

\subsection{Heterogeneous Ocean Representation and Multiscale Spatiotemporal Learning}
Ocean observations differ in their spatial resolution, sampling characteristics, and physical information, motivating the development of methods that can effectively integrate complementary data sources. Thiria et al. combined sea surface height (SSH) and sea surface temperature (SST) observations to reconstruct mesoscale ocean currents through CNN-based downscaling \cite{thiria2023downscaling}. Wang et al. proposed ODF-Net, which fuses in situ measurements, satellite observations, and three-dimensional model and reanalysis data to reconstruct ocean temperature fields \cite{wang2023odfnet}. Champenois and Sapsis developed a machine learning framework that combines historical reanalysis with real-time satellite and buoy measurements for regional four-dimensional ocean temperature reconstruction \cite{champenois2024ocean}. Under sparse observation conditions, Lei et al. investigated multimodal fusion-based spatiotemporal incremental learning for ocean environment perception \cite{lei2024multimodal}. Collectively, these studies explore ocean data fusion through spatial downscaling, multi-source field reconstruction, and multimodal spatiotemporal learning.

Existing ocean spatiotemporal modeling methods primarily employ hierarchical feature extraction, attention mechanisms, and multiscale feature fusion to characterize spatial dependencies and temporal variations in ocean variables. Zhao et al. proposed HSSSA, a hierarchical encoder-decoder network that integrates spatial and temporal self-attention for multiscale SST forecasting \cite{zhao2024hsssa}. Bai et al. developed MUSTAN, which progressively fuses features across spatial scales and applies spatial and temporal attention to predict SST variations \cite{bai2025mustan}. Pan et al. introduced an effective attention model for global SST prediction at daily and monthly temporal resolutions, explicitly modeling spatiotemporal correlations in historical observations \cite{pan2024eam}. Tao et al. investigated multiscale spatial features and temporal attention for El Niño-Southern Oscillation (ENSO) prediction, demonstrating the use of these mechanisms in modeling large-scale ocean variability \cite{tao2024enso}. These studies illustrate complementary architectural strategies for capturing ocean dynamics across spatial scales, temporal horizons, and physical processes.
\section{Method}
\label{sec:method}

\subsection{Overall Architecture}
\begin{figure}[t]
    \centering
    \includegraphics[width=0.85\linewidth]{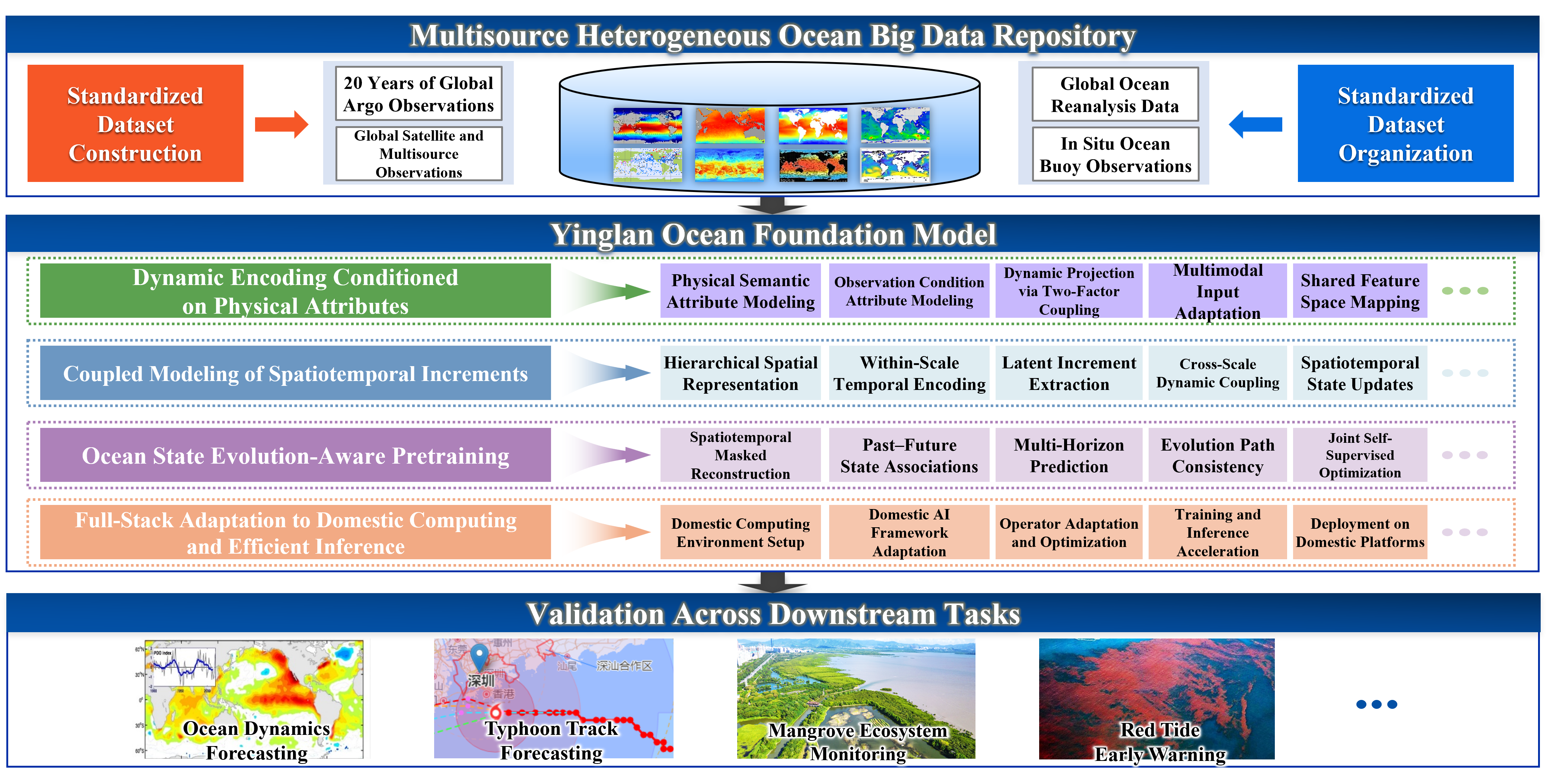}
    \caption{Overall architecture of YingLan.}
    \label{fig:framework}
\end{figure}

We propose YingLan, a multimodal ocean foundation model designed for unified representation learning from heterogeneous ocean observations. The overall architecture is illustrated in Fig.~\ref{fig:framework}. YingLan leverages large-scale satellite remote sensing imagery, gridded oceanographic fields, and buoy time-series observations to address heterogeneity in physical semantics, data structures, and spatiotemporal resolutions across diverse observational sources. The model aims to learn generalizable ocean representations that capture both shared cross-modal information and ocean-specific spatiotemporal dynamics. Its architecture comprises three core components: Physics-Attribute-Conditioned Dynamic Encoding, Spatiotemporal Increment Coupling (STIC), and Ocean-State Evolution-Aware Pretraining.

YingLan first employs Physics-Attribute-Conditioned Dynamic Encoding to construct adaptive feature projections conditioned on the physical semantics and observational characteristics of individual variables. Modality-specific input adapters are used to map ocean observations with different data structures into a shared feature space. The resulting representations are subsequently processed by the STIC module, which integrates multiscale spatial representations with scale-specific temporal modeling. By coupling latent temporal increments across spatial scales, STIC captures dynamic interactions between local ocean structures and large-scale environmental contexts, producing unified spatiotemporal representations.

During pretraining, YingLan jointly optimizes masked observation reconstruction, horizon-conditioned latent increment prediction, and temporal compositional consistency. These complementary objectives enable the model to capture structural dependencies among heterogeneous observations while learning ocean-state variations across multiple temporal horizons. Following pretraining, the shared encoder is adapted to regional ocean monitoring and prediction tasks through task-specific prediction heads and parameter fine-tuning, allowing the transferability of YingLan's pretrained representations to be evaluated in regional marine environments.

\subsection{Physics-Attribute-Conditioned Dynamic Encoding}

Satellite imagery, oceanographic fields, and buoy observations characterize the marine environment from different observational perspectives. Their heterogeneity arises not only from differences in data structures and sampling schemes, but also from the complex  relationships between physical-variable semantics and observation mechanisms. Observations from different sources may represent the same physical variable under different measurement conditions, whereas a single observation platform may provide measurements with distinct physical meanings. Therefore, effective multimodal representation learning requires more than mapping heterogeneous inputs into a common feature space; the encoding process must explicitly capture the relationships between physical semantics and observation conditions.

To address this issue, we propose a physics-attribute-conditioned dynamic embedding mechanism that incorporates the physical semantics and observation conditions of ocean observations into a unified parameter-generation process. The key idea is to formulate the input projection as a learnable operator conditioned on observation attributes, rather than as a fixed mapping predetermined for a specific channel configuration. By separately representing variable semantics and observation conditions and explicitly modeling their coupling, the proposed mechanism aims to distinguish the shared characteristics of different ocean variables from the variations induced by their observation conditions, thereby enabling condition-adaptive representations of heterogeneous ocean observations.

For the $c$-th observation channel, let $\mathbf{m}_c^{s}$ and $\mathbf{m}_c^{o}$ denote its physical-semantic attributes and observation-condition attributes, respectively. The former describes the physical meaning of the observed variable and its associated properties, whereas the latter characterizes measurement conditions, including sensor configuration, observation depth, and spatial resolution, where applicable. These two types of attributes are independently mapped into latent spaces using learnable encoding functions:
\begin{equation}
\mathbf{a}_c^{s}
=
g_{\phi_s}\left(\mathbf{m}_c^{s}\right),
\qquad
\mathbf{a}_c^{o}
=
g_{\phi_o}\left(\mathbf{m}_c^{o}\right).
\end{equation}
where $\mathbf{a}_c^{s}$ and $\mathbf{a}_c^{o}$ denote the physical-semantic representation and observation-condition representation, respectively. For satellite spectral channels, the attributes may include wavelength or spectral-response characteristics. For oceanographic fields, they primarily describe variable categories, physical dimensions, and other relevant physical properties. This factorized attribute representation enables the encoding process to distinguish the physical meaning of an observed variable from the conditions under which it is measured, providing explicit conditioning information for dynamic parameter generation.

To model the joint influence of physical semantics and observation conditions on the input projection, we define the dynamic embedding operator for the $c$-th observation channel as
\begin{equation}
\mathbf{W}_c
=
\mathcal{H}_{\theta}
\left(
\mathbf{a}_c^{s},
\mathbf{a}_c^{o}
\right),
\end{equation}
where $\mathcal{H}_{\theta}$ denotes a learnable dynamic parameter-generation function. To explicitly characterize the coupling between the two types of attributes, we adopt a bilinear factor-coupling parameterization:
\begin{equation}
\mathbf{W}_c
=
\mathbf{U}
\operatorname{Diag}\left(\boldsymbol{\alpha}_c^{s}\right)
\mathbf{C}
\operatorname{Diag}\left(\boldsymbol{\alpha}_c^{o}\right)
\mathbf{V},
\end{equation}
where the attribute-dependent coefficients are generated as
\begin{equation}
\boldsymbol{\alpha}_c^{s}
=
\operatorname{Softmax}
\left(
f_{\theta_s}\left(\mathbf{a}_c^{s}\right)
\right),
\end{equation}
\begin{equation}
\boldsymbol{\alpha}_c^{o}
=
\operatorname{Softmax}
\left(
f_{\theta_o}\left(\mathbf{a}_c^{o}\right)
\right).
\end{equation}
where $\mathbf{U}$ and $\mathbf{V}$ are learnable projection factors, $\mathbf{C}$ is a learnable matrix that captures interactions between physical-semantic and observation-condition factors, and $f_{\theta_s}$ and $f_{\theta_o}$ are coefficient generation functions. The operator $\operatorname{Diag}(\cdot)$ constructs a diagonal matrix from a vector. This parameterization allows physical semantics and observation conditions to modulate distinct latent factors, while their interaction through $\mathbf{C}$ jointly determines the input projection. As a result, the embedding parameters explicitly reflect the combined effects of different observation-attribute configurations.

For satellite imagery and gridded oceanographic fields, let $\mathbf{p}_{i,c}$ denote the local input patch associated with the $c$-th channel at spatial location $i$. Its embedded representation is computed as
\begin{equation}
\mathbf{z}_i
=
\sum_{c=1}^{C}
\mathbf{W}_c
\operatorname{vec}\left(\mathbf{p}_{i,c}\right)
+
\mathbf{b},
\end{equation}
where $C$ denotes the number of input channels, $\operatorname{vec}(\cdot)$ denotes the vectorization operator, and $\mathbf{b}$ is a learnable bias. For buoy observations, the input representation is constructed according to the structure of multivariate time series, incorporating variable attributes, observation timestamps, and spatial locations. The resulting representations are mapped into a latent space with a feature dimension compatible with that of the spatial observations. Consequently, different modalities retain input representations appropriate to their respective data structures while producing tokens suitable for subsequent multiscale spatiotemporal interaction.

Overall, the proposed method formulates the encoding of heterogeneous ocean observations as a dynamic mapping jointly conditioned on physical semantics and observation conditions. Its central mechanism explicitly captures the coupled influence of these attributes on the embedding parameters. The effectiveness of this design will be evaluated against fixed projections, single-attribute-conditioned encoding, and conventional attribute-concatenation-based parameter generation, with particular emphasis on heterogeneous observation
representation and generalization to unseen combinations of variables and observation conditions.

\subsection{Spatiotemporal Increment Coupling (STIC) Module}

The spatiotemporal evolution of oceanic environments exhibits pronounced multiscale dependencies. Local ocean fronts, eddy boundaries, and fine-scale structures are often characterized by strong spatial heterogeneity, whereas large-scale circulation patterns and regional environmental fields provide broader spatial constraints. These structures differ not only in their spatial characteristics but also in their temporal dynamics, with interactions across scales contributing to the complex evolution of ocean states. Conventional factorized spatiotemporal modeling approaches typically capture spatial correlations and temporal dependencies separately, while multiscale fusion methods primarily emphasize feature aggregation across spatial resolutions. Consequently, the dynamic interactions through which temporal variations at different spatial scales jointly shape ocean-state representations remain insufficiently characterized. To address this limitation, we propose a Spatiotemporal Increment Coupling (STIC) module that integrates multiscale spatial representations with scale-specific temporal dynamics. By treating temporal increments of latent states as carriers of cross-scale information exchange, STIC establishes dynamic coupling between local structural variations and large-scale environmental contexts. Unlike conventional approaches that directly fuse static multiscale features, STIC incorporates cross-scale interactions into the state-update process, enabling historical variations at different scales to jointly contribute to spatiotemporal representation learning.

STIC first organizes the encoded ocean observations into hierarchical spatial representations, enabling ocean structures with different spatial extents to be characterized at their corresponding resolutions. Let $\mathbf{Y}_{1:T}=\{\mathbf{Y}_1,\ldots,\mathbf{Y}_T\}$ denote the temporal feature sequence produced by the Physics-Attribute-Conditioned Dynamic Encoding module, where $\mathbf{Y}_t\in\mathbb{R}^{N\times d}$ represents the spatial token embeddings at time step $t$. Here, $T$ denotes the sequence length, while $N$ and $d$ represent the number of spatial tokens and the feature dimension, respectively. Hierarchical spatial features are constructed through scale-dependent transformations:

\begin{equation}
\mathbf{L}_t^{(\ell)}
=
\mathcal{P}_{\ell}
\left(
\mathbf{Y}_t
\right),
\qquad
\ell=0,\ldots,L-1.
\end{equation}
where $\mathbf{L}_t^{(\ell)}\in\mathbb{R}^{N_\ell\times d}$ denotes the feature matrix at spatial scale $\ell$ and time step $t$, $\mathcal{P}_{\ell}$ represents the corresponding spatial scale transformation operator, $N_\ell$ is the number of tokens at that scale, and $L$ denotes the total number of spatial scales. By organizing features at different spatial granularities, the hierarchy accommodates localized ocean fronts, eddy boundaries, and large-scale environmental contexts within their respective spatial extents.

To capture historical variations at different spatial scales, STIC employs scale-specific causal temporal encoders to model temporal dependencies within each spatial hierarchy:

\begin{equation}
\mathbf{B}_t^{(\ell)}
=
\mathcal{T}_{\ell}
\left(
\mathbf{L}_{1:t}^{(\ell)}
\right).
\end{equation}
where $\mathbf{L}_{1:t}^{(\ell)}=\{\mathbf{L}_1^{(\ell)},\ldots,\mathbf{L}_t^{(\ell)}\}$ denotes the historical feature sequence at scale $\ell$ up to time step $t$, and $\mathcal{T}_{\ell}$ is the corresponding causal temporal encoding operator. The resulting latent state $\mathbf{B}_t^{(\ell)}\in\mathbb{R}^{N_\ell\times d}$ integrates historical information at the specified spatial resolution. The causal constraint ensures that each state depends exclusively on observations available up to the current time step. Meanwhile, scale-specific temporal modeling preserves the distinct temporal dependencies of localized ocean structures and broader environmental contexts, avoiding the need to force features at different spatial granularities into a single temporal representation.

Although hierarchical temporal states capture historical dependencies within individual scales, directly aggregating these states does not explicitly distinguish relatively persistent spatial features from temporally varying components. STIC therefore extracts latent temporal increments from scale-specific states and uses them as the basis for cross-scale information transfer:

\begin{equation}
\mathbf{J}_t^{(\ell)}
=
\frac{
\mathbf{B}_t^{(\ell)}
-
\mathbf{B}_{t-1}^{(\ell)}
}{
\delta_t
}.
\end{equation}
where $\mathbf{J}_t^{(\ell)}\in\mathbb{R}^{N_\ell\times d}$ represents the latent temporal increment at scale $\ell$ and time step $t$, $\mathbf{B}_{t-1}^{(\ell)}$ denotes the latent state at the preceding observation time, and $\delta_t>0$ is the actual elapsed time between consecutive observations. The increment characterizes temporal variations in the latent representation rather than the physical time derivative of an observed oceanographic variable. Compared with directly fusing complete scale-specific states, increment-based interaction emphasizes the temporally varying information that links ocean structures across spatial resolutions.

Cross-scale interactions are established through spatially aligned increment transfer. For a target scale $\ell$, latent temporal increments from neighboring scales are mapped to the target spatial resolution and aggregated into a cross-scale dynamic representation:

\begin{equation}
\mathbf{M}_t^{(\ell)}
=
\sum_{j\in\mathcal{N}(\ell)}
\mathcal{A}_{j\rightarrow\ell}
\left(
\mathbf{J}_t^{(j)}
\right).
\end{equation}
where $\mathbf{M}_t^{(\ell)}\in\mathbb{R}^{N_\ell\times d}$ denotes the cross-scale dynamic feature matrix received at the target scale. $\mathcal{N}(\ell)$ is the set of neighboring scale indices, with $j$ representing a source scale. The term $\mathbf{J}_t^{(j)}\in\mathbb{R}^{N_j\times d}$ denotes the latent temporal increment at the source scale, while $\mathcal{A}_{j\rightarrow\ell}:\mathbb{R}^{N_j\times d}\rightarrow\mathbb{R}^{N_\ell\times d}$ represents the spatial mapping operator from scale $j$ to scale $\ell$. By establishing spatial correspondence across different resolutions, this operation allows historical variations in localized structures to contribute to coarse-scale environmental representations, while changes in large-scale backgrounds can be propagated to finer-scale features. Since the transferred information is derived exclusively from current and historical observations, STIC performs causal spatiotemporal encoding without relying on future-state supervision.

To incorporate cross-scale increments directly into target-scale feature transformations, STIC employs a bilinear coupling operator that performs multiplicative interactions between the target-scale historical state and the transferred dynamic information:

\begin{equation}
\begin{aligned}
\mathbf{N}_t^{(\ell)}
=
\mathbf{B}_t^{(\ell)}
+
\Big[
&
\left(
\mathbf{B}_t^{(\ell)}
\boldsymbol{\Theta}_{\mathrm{b}}^{(\ell)}
\right)
\odot
\\
&
\left(
\mathbf{M}_t^{(\ell)}
\boldsymbol{\Theta}_{\mathrm{m}}^{(\ell)}
\right)
\Big]
\boldsymbol{\Theta}_{\mathrm{o}}^{(\ell)}.
\end{aligned}
\end{equation}
where $\mathbf{N}_t^{(\ell)}\in\mathbb{R}^{N_\ell\times d}$ denotes the updated latent state after cross-scale coupling. The learnable projection matrices $\boldsymbol{\Theta}_{\mathrm{b}}^{(\ell)}\in\mathbb{R}^{d\times r}$ and $\boldsymbol{\Theta}_{\mathrm{m}}^{(\ell)}\in\mathbb{R}^{d\times r}$ project the historical state and cross-scale dynamic information into a common interaction space, respectively, while $\boldsymbol{\Theta}_{\mathrm{o}}^{(\ell)}\in\mathbb{R}^{r\times d}$ denotes the output projection matrix. Here, $r$ is the coupling feature dimension, and $\odot$ denotes element-wise multiplication. The bilinear interaction makes the contribution of transferred dynamic information dependent on the historical state at the target scale, thereby capturing conditional relationships between the two representations. Unlike direct concatenation or linear addition, this operator allows cross-scale temporal variations to participate explicitly in latent state transformation rather than serving merely as supplementary feature components.

Following cross-scale coupling, the hierarchical features are reconstructed at the original spatial resolution to produce the final spatiotemporal representation:

\begin{equation}
\mathbf{E}_t
=
\mathbf{Y}_t
+
\sum_{\ell=0}^{L-1}
\mathcal{R}_{\ell}
\left(
\mathbf{N}_t^{(\ell)}
\right).
\end{equation}
where $\mathbf{E}_t\in\mathbb{R}^{N\times d}$ represents the output spatiotemporal feature matrix of STIC, and $\mathcal{R}_{\ell}:\mathbb{R}^{N_\ell\times d}\rightarrow\mathbb{R}^{N\times d}$ denotes the spatial reconstruction operator at scale $\ell$. The input representation $\mathbf{Y}_t$ is retained through a residual connection to preserve the original spatial information. By reconstructing scale-specific dynamic features into a shared spatial resolution, STIC integrates localized ocean structures, large-scale environmental contexts, and their cross-scale dynamic interactions within a unified representation. Through historical increment-driven cross-scale state coupling, the module provides spatiotemporal features for subsequent self-supervised pretraining and downstream task adaptation.

\subsection{Ocean-State Evolution-Aware Pretraining}
Learning transferable representations from heterogeneous ocean observations requires not only the reconstruction of incomplete measurements but also the exploitation of temporal relationships between historical observations and subsequent ocean-state changes. Although masked reconstruction encourages the model to capture observational structures and cross-variable dependencies, it provides limited direct supervision for relating historical ocean states to their future variations. Moreover, independently optimizing predictions over different temporal horizons does not necessarily ensure coherent relationships among the resulting latent transitions. To address these limitations, we incorporate horizon-conditioned latent state prediction and temporal compositional consistency into the self-supervised pretraining objective. Unlike STIC, which models interactions among historical spatiotemporal features within the encoder, this learning strategy exploits observed future states as self-supervisory signals, encouraging the shared encoder to learn ocean representations that support latent state prediction across multiple temporal horizons.

Pretraining employs a dual-branch architecture consisting of an online encoder and a momentum target encoder. The online branch processes randomly masked historical observations through the shared encoding backbone, whereas the target branch extracts supervisory representations from unmasked observations at the corresponding target time steps. The target encoder parameters are updated using an exponential moving average of the online encoder parameters. Considering the variability in spatial coverage and data completeness among heterogeneous oceanographic observations, masked reconstruction is retained as an auxiliary learning objective to preserve spatial structures and cross-variable dependencies:

\begin{equation}
\mathcal{L}_{\mathrm{rec}}
=
\frac{1}{|\Omega|}
\sum_{(u,v)\in\Omega}
\left\|
\widehat{\mathbf{q}}_{u,v}
-
\mathbf{q}_{u,v}
\right\|_2^2.
\end{equation}

Here, $\Omega$ denotes the set of valid masked observation units, while $\mathbf{q}_{u,v}$ and $\widehat{\mathbf{q}}_{u,v}$ represent the normalized ground-truth observation vector and its reconstructed counterpart, respectively. The reconstruction objective preserves observational information, whereas the evolution-aware learning objectives provide additional dynamic supervision beyond instantaneous data recovery.

To establish explicit relationships between historical representations and future ocean-state variations, we employ a shared horizon-conditioned predictor to learn latent state transitions over different prediction intervals:

\begin{equation}
\widehat{\mathbf{D}}_{t,h}
=
f_{\varphi}
\left(
\mathbf{H}_t,
\mathbf{e}_h
\right).
\end{equation}

\begin{equation}
\mathcal{L}_{\mathrm{dyn}}
=
\mathbb{E}_{t,h}
\left[
\left\|
\widehat{\mathbf{D}}_{t,h}
-
\operatorname{sg}
\left(
\overline{\mathbf{H}}_{t+h}
-
\overline{\mathbf{H}}_t
\right)
\right\|_F^2
\right].
\end{equation}

Here, $\mathbf{H}_t$ and $\overline{\mathbf{H}}_t$ denote the latent representations produced by the online encoder and the momentum target encoder, respectively. The vector $\mathbf{e}_h$ encodes the prediction horizon $h$, $\operatorname{sg}(\cdot)$ denotes the stop-gradient operation, and $f_{\varphi}$ represents the shared latent transition predictor. Unlike the historical temporal increments employed internally by STIC, $\widehat{\mathbf{D}}_{t,h}$ represents the predicted latent increment from the current state toward a future target state. By jointly supervising predictions across multiple temporal horizons, the model is encouraged to retain information relevant to subsequent ocean-state changes beyond that captured through observational reconstruction alone.

Latent state transitions over different prediction horizons are further related through a temporal compositional consistency constraint. Given the same historical state, the latent increment predicted directly over a longer interval is encouraged to remain approximately consistent with the cumulative increment obtained through two successive shorter-horizon predictions:

\begin{equation}
\begin{aligned}
\mathcal{L}_{\mathrm{con}}
=
\mathbb{E}_{t,h_1,h_2}
\Big[
\big\|
&
\widehat{\mathbf{D}}_{t,h_1+h_2}
-
\widehat{\mathbf{D}}_{t,h_1}
\\
&
-
f_{\varphi}
\left(
\mathbf{H}_t+
\widehat{\mathbf{D}}_{t,h_1},
\mathbf{e}_{h_2}
\right)
\big\|_F^2
\Big].
\end{aligned}
\end{equation}

Here, $h_1$ and $h_2$ denote the prediction horizons of two successive stages, respectively. The term $\widehat{\mathbf{D}}_{t,h_1+h_2}$ represents the cumulative latent increment predicted directly over the combined horizon, while $\mathbf{H}_t+\widehat{\mathbf{D}}_{t,h_1}$ denotes the intermediate latent state predicted after the first stage. The expectation $\mathbb{E}_{t,h_1,h_2}[\cdot]$ is taken over sampled initial time steps and valid combinations of prediction horizons. This soft consistency constraint relates different prediction trajectories through a shared latent transition mapping without requiring ocean-state evolution to strictly satisfy a prescribed physical governing equation.

The overall pretraining objective is formulated as

\begin{equation}
\mathcal{L}_{\mathrm{pre}}
=
\lambda_{\mathrm{rec}}\mathcal{L}_{\mathrm{rec}}
+
\lambda_{\mathrm{dyn}}\mathcal{L}_{\mathrm{dyn}}
+
\lambda_{\mathrm{con}}\mathcal{L}_{\mathrm{con}}.
\end{equation}

Here, $\lambda_{\mathrm{rec}}$, $\lambda_{\mathrm{dyn}}$, and $\lambda_{\mathrm{con}}$ are weighting coefficients that balance observational reconstruction, future latent state prediction, and temporal compositional consistency, respectively. By incorporating future-observation-based self-supervision, the joint pretraining objective further encourages the shared encoder to capture representations informative of ocean-state variations, providing transferable features for downstream tasks such as coastal ocean monitoring and prediction.

\section{Experiments}
\label{sec:experiments}

\subsection{Datasets}
\subsubsection{Pretraining Datasets}
To support large-scale self-supervised pretraining of YingLan on heterogeneous ocean observations, we established a comprehensive oceanographic data repository integrating in situ observations, satellite remote sensing products, and oceanic and atmospheric reanalysis datasets. The repository incorporates ocean temperature and salinity observations and gridded products from Argo profiling floats, the World Ocean Database (WOD), and EN4; satellite-derived sea surface temperature products from MODIS, AMSR-2, and GHRSST; and oceanic and atmospheric reanalysis datasets from GLORYS, CORA, and NCEP-NCAR. These datasets encompass a wide range of oceanographic and environmental variables, including sea surface temperature, salinity, sea surface height, and surface wind fields. The heterogeneous sources provide complementary data representations, including profiling-float observation sequences, multitemporal satellite imagery, and temporally resolved gridded oceanographic fields, with diverse sampling characteristics. The repository currently contains more than 6 TB of data, spanning multiple observation types, extended temporal periods, and broad geographic coverage, thereby providing a substantial data foundation for unified representation learning and self-supervised pretraining of YingLan across heterogeneous ocean observations.

\subsubsection{Downstream Evaluation Datasets}
To evaluate the transferability of YingLan's pretrained representations to regional marine environments, we construct downstream evaluation datasets using heterogeneous observational data collected from Shenzhen and its adjacent coastal waters. These datasets cover a range of application scenarios, including ocean environmental forecasting, typhoon prediction, coastal ecosystem monitoring, and marine hazard early warning. The data comprise regional gridded oceanographic fields, ocean buoy observations and hydrometeorological time series, multitemporal satellite imagery, unmanned aerial vehicle (UAV) imagery, and associated in situ survey records. Together, these sources support diverse learning tasks, including continuous environmental variable forecasting, time-series regression, remote sensing semantic segmentation, and regional change detection. By evaluating YingLan across regional tasks with distinct data structures and prediction targets, we assess the transferability of its pretrained representations to ocean dynamics characterization, complex coastal environment recognition, and coastal ecosystem change analysis.

\subsection{Experimental Setup}
\subsubsection{Baselines and Evaluation Metrics}
To evaluate the downstream forecasting performance of YingLan,
we compare it against three ocean forecasting models:
LangYa \cite{yang2026langya}, XiHe \cite{wang2024xihedatadrivenmodelglobal}, and GloNet2 \cite{mercator2026glonet2}.
For the seven-day sea surface salinity (SSS) forecasting task,
we report the average root mean square error (RMSE)
across all seven forecast lead times, expressed in
practical salinity units (PSU).
For additional downstream tasks and transfer learning experiments,
evaluation metrics are selected according to the target
variables and task characteristics.
All methods are evaluated using consistent test samples,
spatial grids, and valid-ocean masks to ensure
fair and comparable performance assessments.

\subsubsection{Implementation Details}
YingLan is implemented using the MindSpore deep learning
framework and trained on Huawei Ascend 910B Neural
Processing Units (NPUs).
Model training is conducted within the Ascend computing
environment, leveraging MindSpore's automatic differentiation
and tensor computation capabilities for parameter optimization.
This implementation enables YingLan to be trained on
domestic AI computing infrastructure and provides a
technical foundation for subsequent deployment in
practical ocean monitoring applications.

\subsection{Downstream Task Performance}

We evaluate YingLan on a 7-day sea surface salinity
(SSS) forecasting task to assess its predictive
capability in modeling ocean surface conditions.
The forecasting performance is measured using
the average root mean square error (RMSE) across
seven forecast lead days, with lower values
indicating higher prediction accuracy.

\begin{figure}[t]
    \centering
    \includegraphics[width=0.85\linewidth]{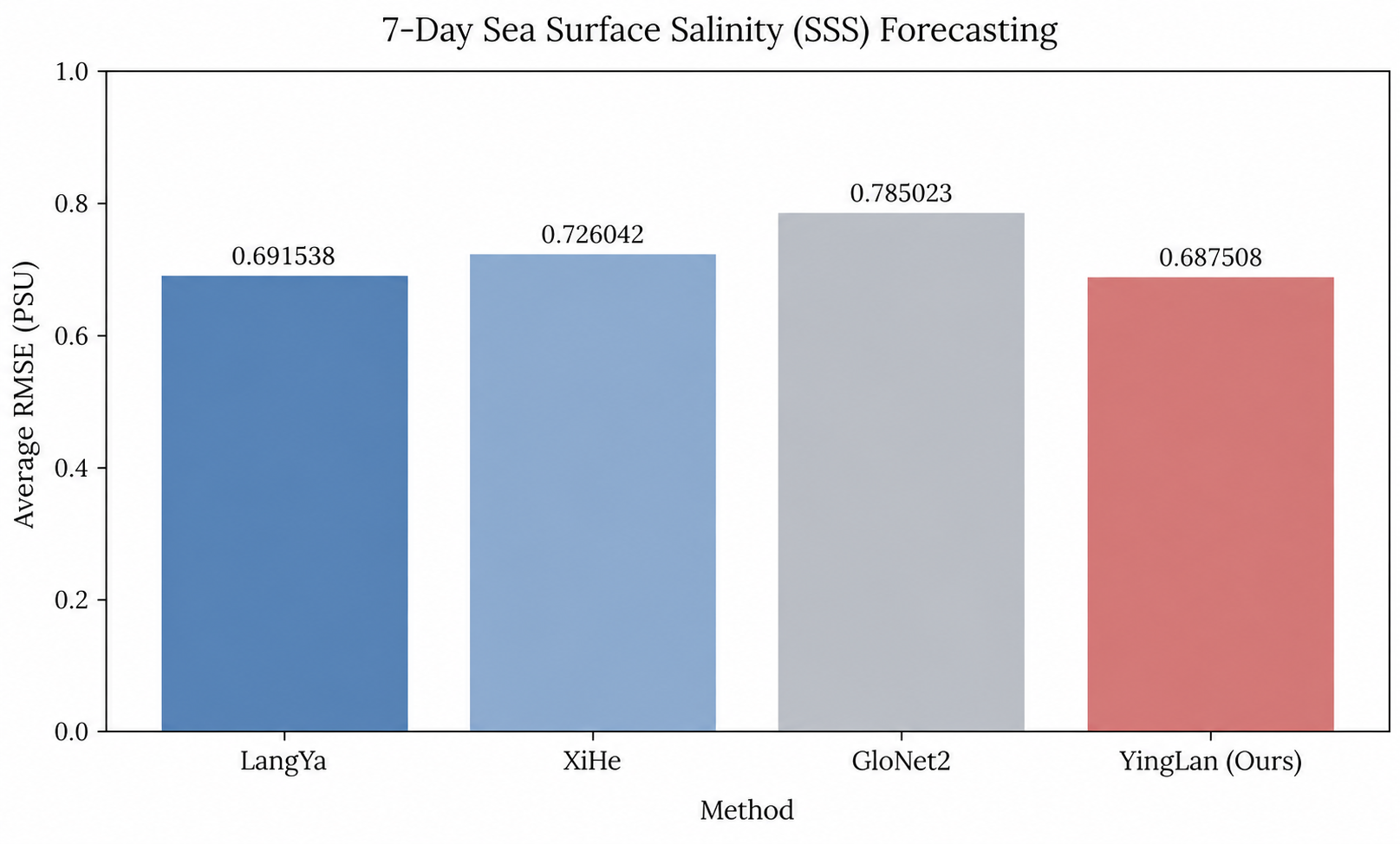}
    \caption{Comparison of the average RMSE for 7-day sea surface salinity (SSS) forecasting. Lower values indicate better forecasting performance.}
    \label{fig:sss_rmse_comparison}
\end{figure}

As shown in Fig.~\ref{fig:sss_rmse_comparison},
YingLan achieves the lowest average RMSE of
0.687508 PSU among the compared methods,
outperforming LangYa (0.691538 PSU),
XiHe (0.726042 PSU), and GloNet2 (0.785023 PSU).
Specifically, YingLan reduces the average RMSE
by 0.58\%, 5.31\%, and 12.42\% compared with
LangYa, XiHe, and GloNet2, respectively.
These results indicate that YingLan demonstrates desirable performance in multi-day sea surface salinity forecasting, highlighting its potential for predicting the state of the ocean surface.

\subsection{Results analysis of ablation experiment}
To investigate the contributions of the key components
of YingLan, we construct three ablation variants by
individually removing the physics-attribute-conditioned
dynamic encoding mechanism, the Spatiotemporal
Increment Coupling (STIC) module, and the
ocean-state evolution-aware pretraining strategy.
All variants are evaluated on the seven-day sea
surface temperature (SST) forecasting task in
Shenzhen coastal waters, using the average RMSE
across seven forecast lead times as the evaluation
metric. The corresponding ablation results are
presented in Table~\ref{tab:ablation_sst}.

\begin{table}[t]
\centering
\caption{Illustrative ablation study of YingLan
on seven-day SST forecasting in Shenzhen coastal
waters. All values are hypothetical placeholders.}
\label{tab:ablation_sst}
\begin{tabular}{lc}
\toprule
Model Variant & Avg. RMSE ($^\circ$C) $\downarrow$ \\
\midrule
w/o Dynamic Encoding            & 0.536 \\
w/o STIC                        & 0.548 \\
w/o Evolution-aware Pretraining & 0.521 \\
\midrule
\textbf{Full YingLan (Ours)}     & \textbf{0.500} \\
\bottomrule
\end{tabular}
\end{table}

\section{Conclusion}
\label{sec:conclusion}

This work investigates unified representation learning from heterogeneous ocean observations and presents YingLan as a multimodal foundation model for ocean environmental modeling. By incorporating physical semantics and observation conditions into adaptive feature encoding, YingLan provides a flexible representation mechanism for ocean variables with diverse observational characteristics. Its multiscale spatiotemporal architecture further integrates temporal variations across spatial hierarchies, while the joint pretraining objectives encourage the learning of both observational structures and ocean-state dynamics. These designs establish a unified modeling framework for integrating satellite imagery, gridded oceanographic fields, and buoy time-series observations. To evaluate its applicability to regional ocean forecasting, we conduct a downstream seven-day sea surface salinity (SSS) forecasting task in the Shenzhen coastal waters. YingLan achieves an average RMSE of 0.687508 PSU, yielding the lowest reported error among the evaluated methods. Beyond predictive performance, YingLan is implemented using MindSpore and trained on Huawei Ascend 910B NPUs, demonstrating the feasibility of developing and training multimodal ocean foundation models within a domestic AI computing environment. Overall, this study provides a methodological basis for learning transferable ocean representations while supporting the development of ocean intelligence models on domestic computing infrastructure. Future research will investigate broader downstream task adaptation, cross-regional generalization, and robust modeling under incomplete and heterogeneous ocean observations.

\section*{Acknowledgments}
Optional acknowledgments go here. Remove this section if unused.

\bibliographystyle{unsrtnat}
\bibliography{references}

@article{YUAN2023159741,
title = {Marine environmental monitoring with unmanned vehicle platforms: Present applications and future prospects},
journal = {Science of The Total Environment},
volume = {858},
pages = {159741},
year = {2023},
issn = {0048-9697},
doi = {https://doi.org/10.1016/j.scitotenv.2022.159741},
url = {https://www.sciencedirect.com/science/article/pii/S0048969722068413},
author = {Shuyun Yuan and Ying Li and Fangwen Bao and Haoxiang Xu and Yuping Yang and Qiushi Yan and Shuqiao Zhong and Haoyang Yin and Jiajun Xu and Ziwei Huang and Jian Lin}
}

@ARTICLE{10.3389/fmars.2024.1363863,
    
AUTHOR={von Jackowski, Anabel },
           
TITLE={Navigating the current landscape of ocean observations: an overview from platform infrastructures to networks related to ocean time series},
          
JOURNAL={Frontiers in Marine Science},
          
VOLUME={Volume 11 - 2024},
  
YEAR={2025},
  
URL={https://www.frontiersin.org/journals/marine-science/articles/10.3389/fmars.2024.1363863},
  
DOI={10.3389/fmars.2024.1363863},
  
ISSN={2296-7745}}

@ARTICLE{10.3389/fmars.2023.1085542,
    
AUTHOR={Chang, Inseong  and Kim, Young Ho  and Jin, Hyunkeun  and Park, Young-Gyu  and Pak, Gyundo  and Chang, You-Soon },
           
TITLE={Impact of satellite and regional in-situ profile data assimilation on a high-resolution ocean prediction system in the Northwest Pacific},
          
JOURNAL={Frontiers in Marine Science},
          
VOLUME={Volume 10 - 2023},
  
YEAR={2023},
  
URL={https://www.frontiersin.org/journals/marine-science/articles/10.3389/fmars.2023.1085542},
  
DOI={10.3389/fmars.2023.1085542},
  
ISSN={2296-7745}}

@ARTICLE{11296896,
  author={Szwarcman, Daniela and Roy, Sujit and Fraccaro, Paolo and Gíslason, Þorsteinn Elí and Blumenstiel, Benedikt and Ghosal, Rinki and de Oliveira, Pedro Henrique and de Sousa Almeida, Joao Lucas and Sedona, Rocco and Kang, Yanghui and Chakraborty, Srija and Wang, Sizhe and Gomes, Carlos and Kumar, Ankur and Gaur, Vishal and Truong, Myscon and Godwin, Denys and Khallaghi, Sam and Lee, Hyunho and Hsu, Chia-Yu and Asanjan, Ata Akbari and Mujeci, Besart and Shidham, Disha and Balogun, Rufai Omowunmi and Kolluru, Venkatesh and Keenan, Trevor and Arevalo, Paulo and Li, Wenwen and Alemohammad, Hamed and Olofsson, Pontus and Mayer, Timothy and Hain, Christopher and Kennedy, Robert and Zadrozny, Bianca and Bell, David and Cavallaro, Gabriele and Watson, Campbell and Maskey, Manil and Ramachandran, Rahul and Moreno, Juan Bernabe},
  journal={IEEE Transactions on Geoscience and Remote Sensing}, 
  title={Prithvi-EO-2.0: A Versatile Multitemporal Foundation Model for Earth Observation Applications}, 
  year={2026},
  volume={64},
  number={},
  pages={1-20},
  doi={10.1109/TGRS.2025.3642610}}

@misc{wang2024xihedatadrivenmodelglobal,
      title={XiHe: A Data-Driven Model for Global Ocean Eddy-Resolving Forecasting}, 
      author={Xiang Wang and Renzhi Wang and Ningzi Hu and Pinqiang Wang and Peng Huo and Guihua Wang and Huizan Wang and Senzhang Wang and Junxing Zhu and Jianbo Xu and Jun Yin and Senliang Bao and Ciqiang Luo and Ziqing Zu and Yi Han and Weimin Zhang and Kaijun Ren and Kefeng Deng and Junqiang Song},
      year={2024},
      eprint={2402.02995},
      archivePrefix={arXiv},
      primaryClass={physics.ao-ph},
      url={https://arxiv.org/abs/2402.02995}, 
}

@online{mercator2026glonet2,
  author = {{Mercator Ocean International}},
  title  = {Digital Ocean Week 2026: Advancing Trusted AI for Europe's Digital Ocean},
  year   = {2026},
  url    = {https://www.mercator-ocean.eu/news-en/digital-ocean-week-2026-ai-ocean-forum/},
  note   = {Official description of the GLONET2 ocean forecasting system}
}

@article{cui2025wenhai,
  title   = {Forecasting the eddying ocean with a deep neural network},
  author  = {Cui, Yingzhe and Wu, Ruohan and Zhang, Xiang and
             Zhu, Ziqi and Liu, Bo and Shi, Jun and Chen, Junshi and
             Liu, Hailong and Zhou, Shenghui and Su, Liang and
             Jing, Zhao and An, Hong and Wu, Lixin},
  journal = {Nature Communications},
  volume  = {16},
  number  = {1},
  pages   = {2268},
  year    = {2025},
  doi     = {10.1038/s41467-025-57389-2},
  url     = {https://doi.org/10.1038/s41467-025-57389-2}
}

@misc{xiong2023aigomslargeaidrivenglobal,
      title={AI-GOMS: Large AI-Driven Global Ocean Modeling System}, 
      author={Wei Xiong and Yanfei Xiang and Hao Wu and Shuyi Zhou and Yuze Sun and Muyuan Ma and Xiaomeng Huang},
      year={2023},
      eprint={2308.03152},
      archivePrefix={arXiv},
      primaryClass={physics.ao-ph},
      url={https://arxiv.org/abs/2308.03152}, 
}

@misc{dawson2025sentinel3foundationmodelocean,
      title={A Sentinel-3 foundation model for ocean colour}, 
      author={Geoffrey Dawson and Remy Vandaele and Andrew Taylor and David Moffat and Helen Tamura-Wicks and Sarah Jackson and Rosie Lickorish and Paolo Fraccaro and Hywel Williams and Chunbo Luo and Anne Jones},
      year={2025},
      eprint={2509.21273},
      archivePrefix={arXiv},
      primaryClass={cs.CV},
      url={https://arxiv.org/abs/2509.21273}, 
}

@misc{stamer2026oceanmaefoundationmodelocean,
      title={OceanMAE: A Foundation Model for Ocean Remote Sensing}, 
      author={Viola-Joanna Stamer and Panagiotis Agrafiotis and Behnood Rasti and Begüm Demir},
      year={2026},
      eprint={2604.08171},
      archivePrefix={arXiv},
      primaryClass={cs.CV},
      url={https://arxiv.org/abs/2604.08171}, 
}

@inproceedings{NEURIPS2023_f6ccbf94,
 author = {Johnson, J. Emmanuel and Febvre, Quentin and Gorbunova, Anastasiia and Metref, Sam and Ballarotta, Maxime and Le Sommer, Julien and fablet, ronan},
 booktitle = {Advances in Neural Information Processing Systems},
 doi = {10.52202/075280-3422},
 editor = {A. Oh and T. Naumann and A. Globerson and K. Saenko and M. Hardt and S. Levine},
 pages = {78275--78295},
 publisher = {Curran Associates, Inc.},
 title = {OceanBench: The Sea Surface Height Edition},
 url = {https://proceedings.neurips.cc/paper_files/paper/2023/file/f6ccbf94fa57c2ae372ece91b537574d-Paper-Datasets_and_Benchmarks.pdf},
 volume = {36},
 year = {2023}
}

@misc{cong2023satmaepretrainingtransformerstemporal,
      title={SatMAE: Pre-training Transformers for Temporal and Multi-Spectral Satellite Imagery}, 
      author={Yezhen Cong and Samar Khanna and Chenlin Meng and Patrick Liu and Erik Rozi and Yutong He and Marshall Burke and David B. Lobell and Stefano Ermon},
      year={2023},
      eprint={2207.08051},
      archivePrefix={arXiv},
      primaryClass={cs.CV},
      url={https://arxiv.org/abs/2207.08051}, 
}

@inproceedings{NEURIPS2023_11822e84,
 author = {Fuller, Anthony and Millard, Koreen and Green, James},
 booktitle = {Advances in Neural Information Processing Systems},
 doi = {10.52202/075280-0241},
 editor = {A. Oh and T. Naumann and A. Globerson and K. Saenko and M. Hardt and S. Levine},
 pages = {5506--5538},
 publisher = {Curran Associates, Inc.},
 title = {CROMA: Remote Sensing Representations with Contrastive Radar-Optical Masked Autoencoders},
 url = {https://proceedings.neurips.cc/paper_files/paper/2023/file/11822e84689e631615199db3b75cd0e4-Paper-Conference.pdf},
 volume = {36},
 year = {2023}
}

@misc{nedungadi2024mmearthexploringmultimodalpretext,
      title={MMEarth: Exploring Multi-Modal Pretext Tasks For Geospatial Representation Learning}, 
      author={Vishal Nedungadi and Ankit Kariryaa and Stefan Oehmcke and Serge Belongie and Christian Igel and Nico Lang},
      year={2024},
      eprint={2405.02771},
      archivePrefix={arXiv},
      primaryClass={cs.CV},
      url={https://arxiv.org/abs/2405.02771}, 
}

@article{bodnar2025foundation,
  title   = {A foundation model for the Earth system},
  author  = {Bodnar, Cristian and Bruinsma, Wessel P. and Lucic, Ana
             and Stanley, Megan and Allen, Anna and Brandstetter, Johannes
             and Garvan, Patrick and Riechert, Maik and Weyn, Jonathan A.
             and Dong, Haiyu and Gupta, Jayesh K. and Thambiratnam, Kit
             and Archibald, Alexander T. and Wu, Chun-Chieh
             and Heider, Elizabeth and Welling, Max
             and Turner, Richard E. and Perdikaris, Paris},
  journal = {Nature},
  volume  = {641},
  pages   = {1180--1187},
  year    = {2025},
  doi     = {10.1038/s41586-025-09005-y},
  url     = {https://doi.org/10.1038/s41586-025-09005-y}
}

@article{eeltink2022nonlinear,
  title={Nonlinear wave evolution with data-driven breaking},
  author={Eeltink, D. and Branger, H. and Luneau, C. and He, Y. and Chabchoub, A. and Kasparian, J. and van den Bremer, T. S. and Sapsis, T. P.},
  journal={Nature Communications},
  volume={13},
  number={1},
  pages={2343},
  year={2022},
  publisher={Nature Publishing Group},
  doi={10.1038/s41467-022-30025-z}
}

@article{ham2024partial,
  title={Partial-convolution-implemented generative adversarial network for global oceanic data assimilation},
  author={Ham, Yoo-Geun and Joo, Yong-Sik and Kim, Jeong-Hwan and Lee, Jeong-Gil},
  journal={Nature Machine Intelligence},
  volume={6},
  number={7},
  pages={834--843},
  year={2024},
  publisher={Nature Publishing Group},
  doi={10.1038/s42256-024-00867-x}
}

@article{glaser2025wvnet,
  title   = {WV-Net: A Foundation Model for SAR Ocean Satellite Imagery},
  author  = {Glaser, Yannik and Stopa, Justin E. and Wolniewicz, Linnea M. and Foster, Ralph and Vandemark, Doug and Mouche, Alexis and Chapron, Bertrand and Sadowski, Peter},
  journal = {Artificial Intelligence for the Earth Systems},
  volume  = {4},
  number  = {4},
  pages   = {e250003},
  year    = {2025},
  doi     = {10.1175/AIES-D-25-0003.1}
}

@inproceedings{bi2024oceangpt,
  title     = {{OceanGPT}: A Large Language Model for Ocean Science Tasks},
  author    = {Bi, Zhen and Zhang, Ningyu and Xue, Yida and Ou, Yixin and Ji, Daxiong and Zheng, Guozhou and Chen, Huajun},
  booktitle = {Proceedings of the 62nd Annual Meeting of the Association for Computational Linguistics},
  pages     = {3357--3372},
  year      = {2024},
  publisher = {Association for Computational Linguistics},
  doi       = {10.18653/v1/2024.acl-long.184},
  url       = {https://aclanthology.org/2024.acl-long.184/}
}

@misc{xue2026oceanpile,
  title         = {{OceanPile}: A Large-Scale Multimodal Ocean Corpus for Foundation Models},
  author        = {Xue, Yida and Zhang, Ningyu and Wu, Tingwei and Ma, Zhe and Ji, Daxiong and Wang, Zhao and Zheng, Guozhou and Chen, Huajun},
  year          = {2026},
  eprint        = {2605.00877},
  archivePrefix = {arXiv},
  primaryClass  = {cs.MM},
  url           = {https://arxiv.org/abs/2605.00877}
}

@article{thiria2023downscaling,
  title   = {Downscaling of ocean fields by fusion of heterogeneous observations using Deep Learning algorithms},
  author  = {Thiria, Sylvie and Sorror, Charles and Archambault, Théo and Charantonis, Anastase Alexandre and Béréziat, Dominique and Mejia, Carlos and Molines, Jean-Marc and Crépon, Michel},
  journal = {Ocean Modelling},
  volume  = {182},
  pages   = {102174},
  year    = {2023},
  doi     = {10.1016/j.ocemod.2023.102174}
}

@article{champenois2024ocean,
  title   = {Machine learning framework for the real-time reconstruction of regional 4D ocean temperature fields from historical reanalysis data and real-time satellite and buoy surface measurements},
  author  = {Champenois, Bianca and Sapsis, Themistoklis},
  journal = {Physica D: Nonlinear Phenomena},
  volume  = {459},
  pages   = {134026},
  year    = {2024},
  doi     = {10.1016/j.physd.2023.134026}
}

@article{lei2024multimodal,
  title   = {Multimodal fusion-based spatiotemporal incremental learning for ocean environment perception under sparse observation},
  author  = {Lei, Lei and Huang, Jie and Zhou, Yu},
  journal = {Information Fusion},
  volume  = {108},
  pages   = {102360},
  year    = {2024},
  doi     = {10.1016/j.inffus.2024.102360}
}

@article{zhao2024hsssa,
  title   = {Hierarchical stacked spatiotemporal self-attention network for sea surface temperature forecasting},
  author  = {Zhao, Yuxin and Yang, Dequan and He, Jianxin and Zhu, Kexin and others},
  journal = {Ocean Modelling},
  volume  = {191},
  pages   = {102427},
  year    = {2024},
  doi     = {10.1016/j.ocemod.2024.102427}
}

@article{bai2025mustan,
  title   = {Multiscale Spatio-Temporal Attention Network for Sea Surface Temperature Prediction},
  author  = {Bai, Zhenxiang and Sun, Zhengya and Fan, Bojie and Liu, An-An and Wei, Zhiqiang and Yin, Bo},
  journal = {IEEE Journal of Selected Topics in Applied Earth Observations and Remote Sensing},
  volume  = {18},
  pages   = {5866--5877},
  year    = {2025},
  doi     = {10.1109/JSTARS.2025.3531122}
}

@article{pan2024eam,
  title   = {Effective attention model for global sea surface temperature prediction},
  author  = {Pan, Xinliang and Jiang, Tao and Sun, Weifu and Xie, Jiawei and Wu, Pinzhen and Zhang, Zhen and Cui, Tingwei},
  journal = {Expert Systems with Applications},
  volume  = {254},
  pages   = {124411},
  year    = {2024},
  doi     = {10.1016/j.eswa.2024.124411}
}

@article{tao2024enso,
  title   = {Multi-scale spatial features and temporal attention mechanisms: advancing the accuracy of ENSO prediction},
  author  = {Tao, Shengen and Li, Yanqiu and Gao, Feng and Fan, Hao and Dong, Junyu and Gan, Yanhai},
  journal = {Intelligent Marine Technology and Systems},
  volume  = {2},
  number  = {1},
  pages   = {7},
  year    = {2024},
  doi     = {10.1007/s44295-023-00017-w}
}

@article{wang2023odfnet,
  title   = {Fusion of ocean data from multiple sources using deep learning: Utilizing sea temperature as an example},
  author  = {Wang, Mingqing and Wang, Danni and Xiang, Yanfei and Liang, Yishuang and Xia, Ruixue and Yang, Jinkun and Xu, Fanghua and Huang, Xiaomeng},
  journal = {Frontiers in Marine Science},
  volume  = {10},
  pages   = {1112065},
  year    = {2023},
  doi     = {10.3389/fmars.2023.1112065}
}

@article{yang2026langya,
  title   = {{LangYa}: A Large AI Model for Global Ocean Forecasting},
  author  = {Yang, Nan and Wang, Chong and Zhao, Zimeng and Zhao, Meihua and Zheng, Huiling and Zhang, Bin and Wang, Fan and Li, Xiaofeng},
  journal = {Science Bulletin},
  year    = {2026},
  doi     = {10.1016/j.scib.2026.08.001},
  url     = {https://doi.org/10.1016/j.scib.2026.08.001}
}

\appendix
\section{Additional Details}
\label{app:details}

Put derivations, extra experimental details, and supplemental tables here.
Delete this appendix if it is unnecessary.

\end{document}